%% file: main.tex
\documentclass[table]{bmvc2k}

\usepackage{amssymb}
\usepackage[capitalize]{cleveref}
\crefname{table}{Tab.}{Tabs.}
\usepackage{algpseudocode}
\usepackage{algorithm}
\usepackage{booktabs}

\title{Gromov Wasserstein Optimal Transport for Semantic Correspondences}

\addauthor{Francis Snelgar}{francis.snelgar@anu.edu.au}{1,2}
\addauthor{Stephen Gould}{stephen.gould@anu.edu.au}{1}
\addauthor{Ming Xu}{mingda.xu@anu.edu.au}{1}
\addauthor{Liang Zheng}{liang.zheng@anu.edu.au}{1}
\addauthor{Akshay Asthana}{akshay.asthana@seeingmachines.com}{2}
\addinstitution{
 School of Computing\\
 Australian National University\\
 Canberra, Australia
}
\addinstitution{
 Seeing Machines\\
 Canberra, Australia
}

\runninghead{Snelgar \etal}{GW Optimal Transport for Semantic Correspondences}

\def\eg{\emph{e.g}\bmvaOneDot}

\def\ie{\emph{i.e}\bmvaOneDot}
\def\etal{\emph{et al}\bmvaOneDot}

\definecolor{ourgray}{gray}{0.9}

\begin{document}

\maketitle

\input{tex_0_abstract.tex}
\input{tex_1_intro.tex}
\input{tex_2_related_works.tex}

\input{tex_3_background.tex}

\input{tex_4_method.tex}

\input{tex_5_experiments.tex}
\input{tex_7_conclusion.tex}

\clearpage
\bibliography{main}
\clearpage
\appendix
\input{tex_8_supplementary.tex}
\end{document}

%% file: tex_0_abstract.tex
\begin{abstract}
    Establishing correspondences between image pairs is a long studied problem in computer vision. With recent large-scale foundation models showing strong zero-shot performance on downstream tasks including classification and segmentation, there has been interest in using the internal feature maps of these models for the semantic correspondence task. Recent works observe that features from DINOv2 and Stable Diffusion (SD) are complementary, the former producing accurate but sparse correspondences, while the latter produces spatially consistent correspondences. As a result, current state-of-the-art methods for semantic correspondence involve combining features from both models in an ensemble. While the performance of these methods is impressive, they are computationally expensive, requiring evaluating feature maps from large-scale foundation models. In this work we take a different approach, instead replacing SD features with a superior matching algorithm which is imbued with the desirable spatial consistency property. Specifically, we replace the standard nearest neighbours matching with an optimal transport algorithm that includes a Gromov Wasserstein spatial smoothness prior. We show that we can significantly boost the performance of the DINOv2 baseline, and be competitive and sometimes surpassing state-of-the-art methods using Stable Diffusion features, while being 5--10x more efficient. We make code available at \url{https://github.com/fsnelgar/semantic_matching_gwot}.
\end{abstract}

%% file: tex_1_intro.tex
\section{Introduction}

\begin{figure}
    \centering
    \includegraphics[width=0.9\textwidth]{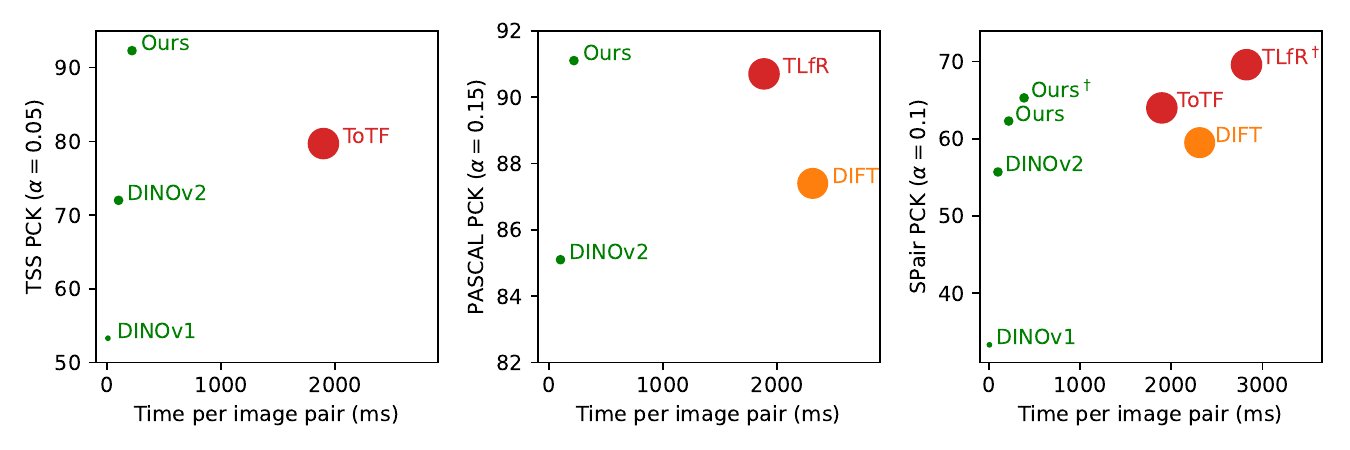}
    \caption{Accuracy and latency trade off for various methods. Our method achieves comparable accuracy while being much faster and requiring less memory. The size of the markers indicates the relative number of parameters in each method, and the colour indicates the model family. Green methods use DINO models, orange methods use Stable Diffusion, and red methods combine features from both. Methods marked with $\dagger$ use ground truth labels to flip keypoints at test time. See \cref{tab:comp_results} for a detailed breakdown.}
    \label{fig:latency}
\end{figure}

In contrast to the traditional counterpart, the \textit{semantic} matching task requires finding correspondences between different instances of the same class (\eg, the ears of two different cats) across images \cite{rocco_convolutional_2019,rocco_end--end_2017,laskar_semantic_2019,cho_cats_2021,cho_cats_2023,schonberger_semantic_2018}. It is an inherently challenging problem in computer vision due to large visual differences between classes, non-rigid objects and change in appearance due to camera pose. The task has numerous interesting applications including visual localisation \cite{li_accurate_2022,schonberger_semantic_2018} semantic and few-shot segmentation \cite{ravi_sam_2024, kang_integrative_2022} and image transfer \cite{tumanyan_disentangling_2023,tumanyan_splicing_2022}. There are several properties that are desirable in a semantic correspondence method. First, correspondences should be unique --- each pixel in the target image should match at most one pixel in the source image. Second, the correspondence map should be locally spatially smooth, a pair of pixels that are sufficiently close in the source image should match to a pair that are similarly close in the target image, but global non rigid deformations mean this property is only desirable locally. Third, a matching algorithm must accommodate objects of different scale as well as background regions which have no semantic correspondences.
Recent methods \cite{tang_emergent_2023,caron_emerging_2021,zhang_telling_2023,zhang_tale_2023,fundel_distillation_2025,luo_diffusion_2023} using pretrained foundation models \cite{caron_emerging_2021,rombach_high-resolution_2022} have identified that the spatial smoothness property can be provided by Stable Diffusion features due to its strong spatial awareness in contrast to DINO features. We instead take a different approach and design a novel matcher based on optimal transport, encoding all properties --- including spatial smoothness --- into the matching algorithm, yielding competitive results with a fraction of the compute and memory. We summarise the speed and performance of zero-shot methods in \cref{fig:latency}, with our method achieving competitive performance at a fraction of the cost.

%% file: tex_2_related_works.tex
\section{Related Works}
\subsection{Semantic Correspondences}
Early semantic correspondence methods followed keypoint matching pipelines with hand-crafted descriptors  \cite{dalal_histograms_2005,rublee_orb_2011,lowe_object_1999} combined with matching algorithms using geometric models \cite{liu_sift_2011,cho_unsupervised_2015}. Cho~\etal~\cite{cho_unsupervised_2015} uses hierarchical object proposals and the Probablistic Hough Transform (PHM) as a geometric model. ProposalFlow~\cite{ham_proposal_2016} extends this work, addressing the limitations of the global matching consensus. SiftFlow~\cite{liu_sift_2011} combines coarse-to-dense SIFT descriptors with a belief propagation layer to encourage spatial smoothness.

Later methods \cite{rocco_convolutional_2019,rocco_end--end_2017,long_convnets_2014,liu_semantic_2020,laskar_semantic_2019} use convolutional neural networks in replacement of traditional descriptors. Long~\etal~\cite{long_convnets_2014} showed features from convolutional networks worked as well as SIFT \cite{lowe_object_1999} descriptors for alignment and classification tasks. Rocco~\etal~\cite{rocco_end--end_2017} uses a parametric geometric layer to train CNNs end-to-end, later replacing it with a RANSAC \cite{fischler_random_1981} inspired layer to allow weakly supervised training. SCOT \cite{liu_semantic_2020} uses a network pretrained on ImageNet \cite{deng_imagenet_2009} with optimal transport \cite{thorpe_introduction_2018} and PHM \cite{cho_unsupervised_2015} post processing.

In recent years a body of work \cite{mariotti_improving_2024, zhang_tale_2023, zhang_telling_2023, tang_emergent_2023,fundel_distillation_2025, luo_diffusion_2023} has focused on using large foundation models including DINO \cite{caron_emerging_2021,oquab_dinov2_2023,darcet_vision_2023} and Stable Diffusion \cite{rombach_high-resolution_2022} due to their powerful zero shot performance on downstream tasks. DIFT~\cite{tang_emergent_2023} use features from Stable Diffusion prompted with the class label, while Zhang~\etal~\cite{zhang_tale_2023,zhang_telling_2023} uses the ODISE \cite{xu_open-vocabulary_2023} pipeline, including an implicit prompt from CLIP \cite{radford_learning_2021}. Several works combine features from several foundation models, either using PCA \cite{zhang_tale_2023} or distillation \cite{fundel_distillation_2025} to reduce the computational burden. Several methods attempt to integrate the global object pose to resolve visual ambiguity, Zhang~\etal~\cite{zhang_telling_2023,fundel_distillation_2025} uses labelled symmetric keypoints while Mariotti~\etal~\cite{mariotti_improving_2024} trains a spherical geometric prior to map DINO features to the unit sphere in object coordinates.

Different to these works, our contribution aims to improve the matching algorithm using features from a single model. We show that we can imbue desirable properties of foundation model features into the matching algorithm itself, while using less compute and memory. Our method is related to SCOT \cite{liu_semantic_2020}, which also uses optimal transport for matching. However, SCOT does not include properties such as spatial smoothness and symmetry within the OT problem. We show including these properties significantly improves matching accuracy.

\subsection{Optimal Transport in Computer Vision}
As our method is based on optimal transport (OT), we provide a brief review of related applications where OT is used for matching or alignment. We also discuss related works where Gromov Wasserstein OT has been used to imbue structure into the matching algorithm.

Optimal transport is a long studied problem ~\cite{monge_memoire_1781, kantorovich_translocation_2006} and there are many variants including partial \cite{sarlin_superglue_2020, chapel_partial_2020} and unbalanced \cite{a_khamis_scalable_2024} transport and the Gromov Wasserstein  \cite{memoli_gromovwasserstein_2011, peyre_gromov-wasserstein_2016} formulation. For an excellent introduction into the problem we refer the reader to Thorpe's notes \cite{thorpe_introduction_2018}. Applications of optimal transport include keypoint matching \cite{sarlin_superglue_2020}, positive-unlabelled learning \cite{chapel_partial_2020} and object detection \cite{plaen_unbalanced_2023}. Others have exploited structure in the problem with Gromov Wasserstein optimal transport for temporal action segmentation \cite{xu_ming_temporally_2024}, graph classification \cite{titouan_optimal_2019} alignment of fMRI data \cite{thual_aligning_2022} and domain adaptation \cite{gu_keypoint-guided_2022}. In this work we apply Gromov Wasserstein optimal transport to the new application of semantic correspondences.

%% file: tex_3_background.tex
\section{Background}
In this section we provide background on optimal transport which forms the basis of our method. First, we introduce the OT problem and how it can be used for matching problems, \eg, \cite{sarlin_superglue_2020}. Next we introduce Gromov Wasserstein OT, which allows incorporating structure such as spatial structure, leveraging recent works \cite{xu_ming_temporally_2024, peyre_gromov-wasserstein_2016}.

\subsection{Kantorovich Optimal Transport}

Optimal transport (OT) ~\cite{monge_memoire_1781} is the problem of comparing two distributions and moving one to the other in the most efficient way possible. For the classic Kantorovich~\cite{kantorovich_translocation_2006} OT with discrete measures $p \in \Delta_n$, $q \in \Delta_m$, where $\Delta_k$ is the $k-1$ dimensional probability simplex, and a (symmetric) ground cost $C \in \mathbb{R}^{n \times m}_{+}$ (\ie, the `cost' of transporting mass from element $p_i$ to $q_i$) the optimal transport map $T^\star$ is the solution to the linear program:
\begin{equation}
    T^{\star} = \operatornamewithlimits{argmin}_{T \in \mathcal{T}}~ \langle C, T \rangle,
    \label{eq:kot}
\end{equation}
where $\mathcal{T}$ is the transport polytope $\mathcal{T} = \{ T \in \mathbb{R}^{n \times m}_{+} : T\mathbf{1}_m = p, T^\top\mathbf{1}_n = q \}$. For the applications in this work we consider $T$ as a soft assignment between elements of $p$ and $q$.

\subsection{Gromov Wasserstein Optimal Transport}

The optimal transport formulation introduced so far require that there exists a ground cost $C$ between $p$ and $q$. Gromov Wasserstein (GW) \cite{memoli_gromovwasserstein_2011, peyre_gromov-wasserstein_2016} instead defines the transport cost between pairwise elements within the same measure.  Given (symmetric) $C^p \in \mathbb{R}^{n \times n}$, $C^q \in \mathbb{R}^{m \times m}$ which defines a distance metric between pairs of elements in $p$ (resp. $q$), the GW optimisation objective is given by
\begin{equation}
     \mathcal{F}_\text{GW}\left(C^p, C^q, T\right) = \sum_{\substack{(i,k) \in [n] \times [n] \\  (j,l) \in [m] \times [m]}} L(C^p_{i,k}, C^q_{j,l})T_{i,j}T_{k,l},
     \label{eq:gw_def}
\end{equation}
where $L$ defines a distance metric between elements of cost matrices $C^p$ and $C^q$. As noted by Peyr\'e~\cite{peyre_gromov-wasserstein_2016} for $L\left(C^p_{i,k},C^q_{j,l}\right) =C^p_{i,k}C^q_{j,l}$, $\mathcal{F}_\text{GW}$ can be computed efficiently using only matrix operations as $\langle C^p T C^{q\top}, T \rangle$.

Kantorovich and GW optimal transport can be combined when there is a structural prior as well as a defined ground cost to form the Fused-GW problem \cite{thual_aligning_2022, titouan_optimal_2019},
\begin{equation}
      T^{\star} = \operatornamewithlimits{argmin}_{T \in \mathcal{T}}~ \langle C, T \rangle + \mathcal{F}_\text{GW}\left(C^p, C^q, T\right).
\end{equation}

%% file: tex_4_method.tex
\section{Optimal Transport for Semantic Correspondences}
\label{sec:desc_method}
In this section we describe our method for semantic correspondences. We briefly introduce the task before detailing how we design an optimal transport based matching algorithm that encodes the properties of the semantic correspondence task.

\subsection{Problem Statement}
Given RGB image pairs $I, \hat{I} \in \mathbb{R}^{H \times W \times 3}$, we wish to find correspondences between pixels in $I, \hat{I}$ which represent semantically `similar' parts. For each image, we assume we have access to the (normalised) feature maps from a foundation model $y = f_\theta \left( I \right) \in \mathbb{R}^{N \times D}$, where $N=\frac{H \times W}{P^2}$ is the spatial size of the feature maps with corresponding patch coordinates $\mathcal{X} \in [1, \frac{W}{P}] \times [1, \frac{H}{P}]$ and size $P$. To establish correspondences, we solve the optimisation problem
\begin{equation}
    T^{\star} = \operatornamewithlimits{argmin}_{T}~ \lambda \langle C, T \rangle + \lambda_\text{gw} \mathcal{F}_\text{GW}\left(C^p, C^q, T\right) + \lambda_\text{sym}\mathcal{F}_\text{sym}\left( T \right) + \lambda_\text{ub} \text{D}_\text{KL}(T^\top\mathbf{1}_N, q),
      \label{eq:keypoint_ot}
\end{equation}
where $p = \frac{1}{N}\mathbf{1}_N$, $q = \frac{1}{N}\mathbf{1}_N$ are measures over the respective sets of coordinates $\mathcal{X}$, $\hat{\mathcal{X}}$ and $\{T \in \mathbb{R}^{N \times N} : T\mathbf{1}_N = p\}$. The correspondence $\hat{x}_j$ for patch $x_i$ is given by $j=\operatorname{argmax} \left(T^\star_{i}\right)$. We explain the exact form and reasoning for the terms in \cref{eq:keypoint_ot} in the following sections.

\subsection{Objective Functions}

\noindent{\textbf{Feature Similarity:}} The cosine similarity between normalised features $y,\hat{y}$ captures the semantic similarity between patches, we therefore use $C= 1- y\hat{y}^\top$ as the ground cost. However feature similarity is spatially noisy, and by itself does not encourage spatial smoothness.

\begin{figure}
    \centering
    \includegraphics[width=0.95\textwidth]{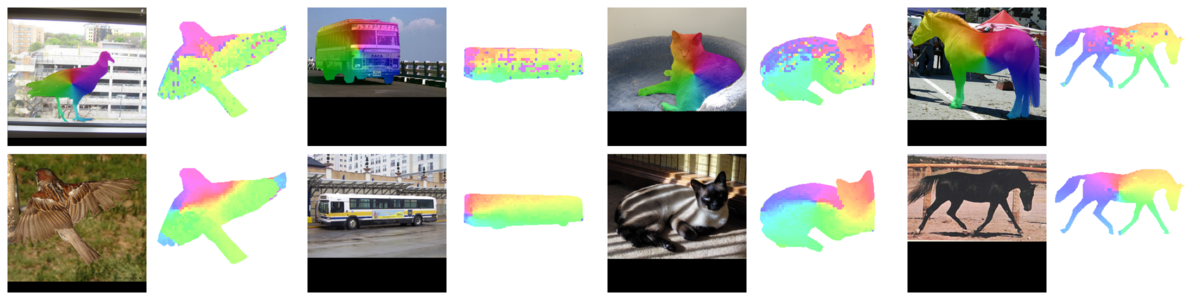}
    \caption{Impact of Gromov Wasserstein optimal transport. Each two-by-two grid contains an example image pair on the left, and  dense correspondences using nearest neighbours and our GW method in the top right and bottom right respectively. Note that correspondences for our method are more spatially consistent.}
    \label{fig:gw_examples}
\end{figure}

\noindent{\textbf{Spatial Smoothness:}} To encourage spatial smoothness we include a Gromov Wasserstein objective $\mathcal{F}_\text{GW}\left(C^p, C^q, T \right) = \langle C^p T C^{q\top}, T \rangle$, with cost matrices
\begin{equation}
    C^p_{i,k} =
    \begin{cases}
        1 ,& || x_i - x_k||_2 < \delta_\text{min} \\
        0, & \text{otherwise}
    \end{cases}
    \qquad
    C^q_{j,l} =
    \begin{cases}
        1 ,& || \hat{x}_j - \hat{x}_l||_2 > \delta_\text{max} \\
        0, & \text{otherwise}.
    \end{cases}
    \label{eq:sem_kp_gw_cost_1}
\end{equation}
Consider two pairs of patches with coordinates $x_i,x_k$ and $\hat{x}_j, \hat{x}_l$ in images $I$ and $\hat{I}$, respectively. The GW objective $\mathcal{F}_\text{GW}$ is non-zero for pairs that are within radius $\delta_\text{min}$ in $I$ and further than radius $\delta_\text{max}$ in $\hat{I}$. This encourages neighbouring patches to match to patches that are similarly close, while allowing for small amounts of local non rigid deformation. This results in matches that are more spatially consistent, as shown in \cref{fig:gw_examples}. Note that since $C^p,C^q$ are sparse, we can implement it using 2D convolutions, avoiding expensive matrix multiplication operations to further reduce compute and memory requirements.

\begin{figure}
    \centering
    \includegraphics[width=0.75\textwidth]{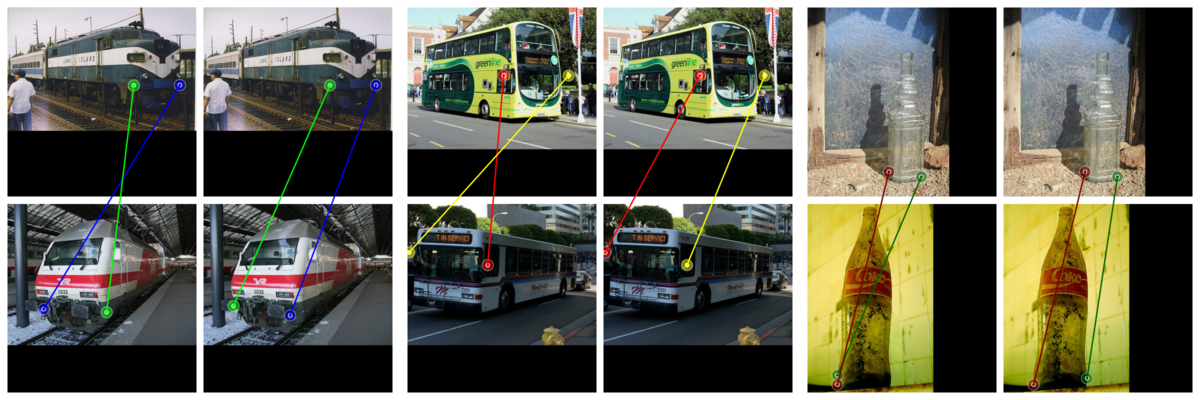}
    \caption{Effect of the symmetry loss. Each two-by-two grid contains matches without the symmetry loss on the left, and with the symmetry loss on the right. Note that with the symmetry loss ordering of the keypoint pair is maintained and matches are more plausible.}
    \label{fig:symmetry_loss_examples}
\end{figure}

\noindent{\textbf{Object Symmetry:}} Similar to Zhang~\etal~\cite{zhang_telling_2023} we note that there often exists local ambiguity in matching between objects, where the global geometry of the object is required to correctly resolve local regions. For example, a `wheel' may be visually similar to multiple other wheels, and the correct match can only be determined by considering the global pose of the vehicle. To address this we introduce a symmetry aware objective. We observe that many objects have a vertical axis of symmetry, and that under moderate changes in pose, the ordering of keypoints around this axis should be consistent across images.  Let $\mathcal{G} = \{u_0, u_1, \ldots, u_M\}$ be the set of $M$ pairs of symmetric keypoints for a given object where $u_m = \left( x_i, x_j \right)$ are the patch coordinates for the pair in $I$. The symmetry object function is
\begin{equation}
    \mathcal{F}_\text{sym}\left(S, T \right) =  \sum_{\substack{(i,k) \in \mathcal{G} \\  (j,l) \in [N]}} -S_{i,j}S_{k,l}T_{i,j}T_{k,l},
\end{equation}
where $S\in \mathbb{R}^{N \times N}$ and $S_{i,j} = \text{sign} (x^{(0)}_i - x^{(0)}_j)$ indicating the relative ordering of patch coordinates horizontally (along the x-axis). While inspired by Gromov Wasserstein, $\mathcal{F}_\text{sym}$ is not a strictly GW as $S$ is not a valid metric space on $\mathcal{X}$ due to $S_{i,j} \neq S_{j,i}$. Nevertheless, it is still a valid objective function and can be minimized in the same problem. We show examples of this in \cref{fig:symmetry_loss_examples}. Without the symmetry loss keypoints are matched incorrectly to opposite sides of the object. With the loss, ordering is maintained and keypoints are matched correctly.

\subsection{Unbalanced Formulation}

\begin{figure}
    \centering
    \includegraphics[width=0.95\textwidth]{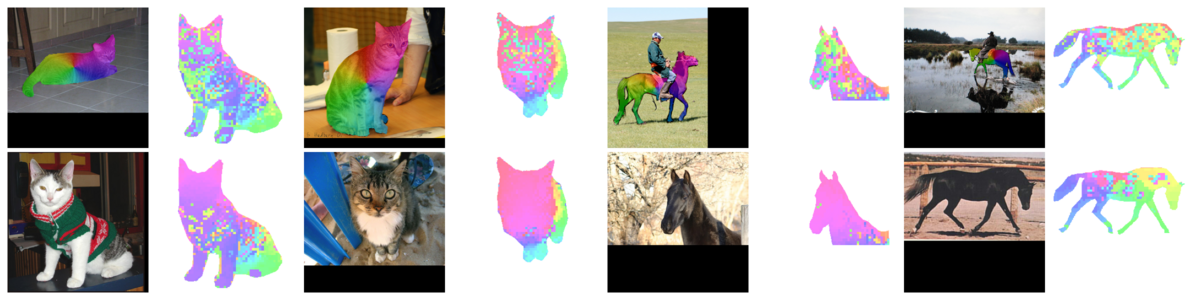}
    \caption{Impact of unbalanced optimal transport. Each two-by-two grid shows the image pair on the left, with results for balanced and partially balanced optimal transport on the top right and bottom right respectively. For objects with large scale differences (first and fourth examples) or occlusion (second and third examples), pixels have unequal importance and the balanced mass assumption doesn't hold. The unbalanced OT results are more plausible.}
    \label{fig:ub_examples}
\end{figure}

A balanced optimal transport problem would create one-to-one correspondences between all patches. However in practice this is not always desirable as object scale may vary significantly, and there may be regions which are not covisible. Therefore we relax the balanced assignment constraint to form a partially unbalanced problem. The first marginal constraint $T\mathbf{1}_N = p$ remains, insuring that a match is found for all patches in $I$, however the second constraint $T^\top\mathbf{1}_N = q$ is replaced with a KL divergence regularisation penalty $\text{D}_\text{KL}(T^\top\mathbf{1}, q)$. We show examples of the impact of the unbalanced formulation in \cref{fig:ub_examples}.

%% file: tex_5_experiments.tex
\section{Experiments}
\label{sec:experiments}
\subsection{Implementation Details}
We follow previous zero-shot methods and use a pre-trained and frozen foundation model as a feature extractor. We use the DINOv2 (ViT-B/16) model with an image resolution of 840-by-840 and extract all tokens except the class token from the last layer as features for matching. For the symmetric aware objective, we use the coordinates of the keypoints in the source image provided in the datasets for the symmetric pairs, defined by Zhang~\etal~\cite{zhang_telling_2023}. We solve the OT problem in \eqref{eq:keypoint_ot} using projected gradient descent with 50 steps. We provide a full breakdown of hyperparameters in the supplementary material.

\subsection{Datasets}
\noindent{\textbf{TSS:}} The TSS dataset \cite{taniai_joint_2016} contains 400 image pairs primarily of vehicles curated from existing FG3D \cite{lin_jointly_2014}, PASCAL \cite{hariharan_semantic_2011} and JODS \cite{rubinstein_unsupervised_2013} datasets. As vehicles are non deformable, methods benefit from having a strong spatial smoothness prior. Dense correspondence labels are provided for the objects of interest.

\noindent{\textbf{PF-PASCAL:}} PF-PASCAL \cite{ham_proposal_2016} contains has 1,351 image pairs across 20 object categories. Keypoints are provided from the 2011 PASCAL \cite{bourdev_poselets_2009} annotations. The dataset is more challenging than TSS, with greater variation in pose and scale.

\noindent{\textbf{SPair-71k:}} SPair \cite{min_spair-71k_2019} is a highly challenging semantic correspondence dataset with large variation in object pose and a higher degree of scene clutter. It is significantly larger than previous datasets, containing 70,958 image pairs of 18 categories with keypoint annotations.

\subsection{Metrics}
We use the standard Percentage of Correct Keypoints (PCK) metric for evaluation. For each keypoint, the predicted correspondence is considered correct if it is within $\alpha \cdot \text{max}(w, h)$ radius of the ground truth match, where $0 \leq \alpha \leq 1$ and $w,h$ are the width and height of the image for TSS and PF-PASCAL, or the bounding box for SPair-71k. We report results using the total number of correct keypoints in a category (or dataset) normalized by the total number of keypoints for the SPair-71k dataset, and normalised per image for TSS and PF-PASCAL datasets following recent prior works \cite{zhang_telling_2023,tang_emergent_2023,zhang_tale_2023,fundel_distillation_2025}.

\begin{table}[t!]
    \centering
    \scriptsize
    \input{tables_tss_pascal_results.tex}
    \caption{Performance on the TSS and  PF-PASCAL datasets. We report \texttt{per-image} PCK results. Results in the top half of the table are for supervised methods, results in the bottom half of the table are for zero-shot methods. Results for our method are coloured in grey. Methods marked with $^\star$  are taken from \cite{zhang_telling_2023}.}
    \label{tab:tss_pascal_results}
\end{table}

\begin{table}
    \centering
    \scriptsize
    \setlength{\tabcolsep}{1.5pt}
    \input{tables_spair_results.tex}
    \caption{Performance on the Spair-71k dataset. Due to inconsistencies in prior works we separate methods based on if they report \texttt{per-keypoint} or \texttt{per-image} results.  Results in the top half of tables are for supervised methods, results in the bottom half of tables are for zero-shot methods. Results marked with $\dagger$ use ground truth annotations to reorder keypoints at test time for the Pose-Align method from \cite{zhang_telling_2023}. Results for our method are coloured in grey.}
    \label{tab:spair_results}
\end{table}

\subsection{Results}

\noindent{\textbf{Comparison to State-of-the-art:} We compare our method with state-of-the-art methods for the TSS and PASCAL datasets in \cref{tab:tss_pascal_results}. Our method improves on state-of-the-art for the TSS dataset, which we attribute to the fact that the dataset contains image pairs with low pose variation and non-deformable objects (\eg, cars, trains, buses) and as such, a strong spatial smoothness prior is particularly beneficial. Our performance on the PASCAL dataset is comparable to existing zero-shot methods but still improves significantly over the nearest neighbour baseline at all thresholds. We also present results for the much more challenging Spair-71k dataset in \cref{tab:spair_results}. Our method lags slightly behind state-of-the art methods that use Stable Diffusion features, particularly in the \texttt{Plant} and \texttt{TV} categories. We attribute this to the strength of the underlying features, the Stable Diffusion (SD) features outperform DINOv2 features by over 20 percent in these categories. While our method significantly closes the gap, it still relies on strong features. We also include results using the Pose-Align technique from Zhang~\etal~\cite{zhang_telling_2023}, showing that it is orthogonal to our method and can further boost performance at the expensive of additional compute.

\begin{table}
    \centering
    \scriptsize
    \input{tables_comp_results.tex}
    \caption{Computational requirements for zero-shot methods. For Stable Diffusion methods we include the VAE encoder and UNet in the parameter count. For DIFT, we include the text encoder as the method uses captions, for ToTF and TlFR we include the visual encoder from CLIP used for implicit captions. Feat.~Ext. is the time for extracting features from an image pair, Match is the time for matching between features.}
    \label{tab:comp_results}
\end{table}
\noindent{\textbf{Computational Performance:}} We compare computational requirements for our method and other zero-shot methods in \cref{tab:comp_results} and \cref{fig:latency}. All results are produced from the official repositories and are the average of 100 runs on an Nvidia RTX-4090 GPU. It is clear that other methods are dominated by feature extraction with the matching latency relatively inconsequential. In comparison our matching latency is higher, but still significantly less than Stable Diffusion. Overall our method requires less computational resources while still being competitive with more expensive state-of-the-art methods.
\begin{table}
    \centering
    \scriptsize
    \input{tables_ablation.tex}

    \caption{Ablation study on the SPair-71k dataset.}
    \label{tab:ablation}
\end{table}

\noindent{\textbf{Ablation Study:}} We perform an ablation study of the different elements of our method in \cref{tab:ablation}. The GW, unbalanced (UB), and symmetry objective (Sym) all provide meaningful improvements over the baseline. The Pose-Align method \cite{zhang_telling_2023} can be applied in addition to our method for further improvement.

\noindent{\textbf{Limitations:}} The GW spatial smoothness term is based on the premise that correspondences are spatially consistent --- at least within a local neighbourhood. However in cases of extreme scale or pose change this assumption does not hold. Recall \cref{eq:sem_kp_gw_cost_1} penalises pairs of patches within a radius of $\delta_\text{min}$ that match to patches outside the radius $\delta_\text{max}$. The first example in \cref{fig:failure_examples} shows a case where our choice of $\delta_\text{min},\delta_\text{max}$ is inconsistent with the extreme scale difference. An interesting direction for future work would be to develop methods for estimation of adaptive filter sizes. Our method excels at removing noisy outliers, however can create smooth regions of incorrect correspondences as the proportion of outliers increases as shown in the second example. The one-to-one correspondence assumption is not always valid as shown in the third example. As we are using patch features at a lower spatial resolution, the small scale of the horse means several keypoints in the horse's head belong to the same patch. Lastly our symmetry assumption only holds for moderate pose variations, the fourth example shows an example of opposite viewpoints where this assumption is not valid.
\begin{figure}
    \centering
    \includegraphics[width=0.95\linewidth]{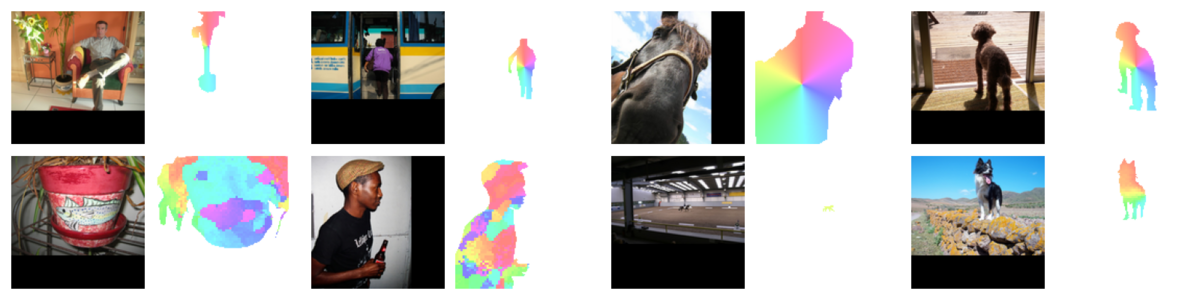}
    \caption{Failure cases of our method. Each two-by-two grid shows dense correspondences for a failure case. First and third examples show failure due to scale, second shows failure due to incorrect GW smoothing, and fourth failure due to invalid symmetry assumption.}
    \label{fig:failure_examples}
\end{figure}

%% file: tables_tss_pascal_results.tex
\begin{tabular}{l|cccc|ccc}
    \toprule
     & \multicolumn{4}{c|}{\textsc{tss pck}~$\alpha_\text{image} = 0.05$} & \multicolumn{3}{c}{\textsc{pascal pck}~$\alpha_\text{image} = k$} \\
    \midrule
    Method                                          & FG3DCar & JODS   & Pascal & Avg.    &   0.05 &    0.1 &   0.15\\
    \midrule
    CATs \cite{cho_cats_2021}                       &    92.1 &   78.9 &   64.2 &   78.4  &   76.8 &   92.7 &   96.5 \\
    CATs++ \cite{cho_cats_2023}                     &   -     &     -  &    -   &   -     &   84.9 &   93.8 &   96.8 \\
    PWarpC-CATs \cite{truong_probabilistic_2022}    &    95.5 &   85.0 &   85.5 &   88.7  &   79.8 &   92.6 &   96.4 \\
    CNNGeo \cite{rocco_convolutional_2019}          &   90.1 &   76.4 &   56.3 &   74.4   &   41.0 &   69.5 &   80.4 \\
    Semantic-GLU-Net \cite{truong_warp_2021}        &   95.3 &   82.2 &   78.2 &   85.2   &   48.3 &   72.5 &   85.1 \\
    \midrule
    SCOT \cite{liu_semantic_2020}                   &    \underline{95.3} &   \underline{81.3} &   57.7 &   78.1  &   63.1 &   \underline{85.4} &   \textbf{92.7} \\
    DINOv1 \cite{zhang_tale_2023}                    &   64.7 &   51.2 &   36.7 &   53.3   &     -  &    -   &   -    \\
    DINOv2$^\star$ \cite{zhang_tale_2023}            &   82.8 &   73.9 &   53.9 &   72.0   &   63.0 &   79.2 &   85.1 \\
    SD \cite{zhang_tale_2023}                        &   93.9 &   69.4 &   57.7 &   77.7   &     -  &    -   &   -    \\
    DIFT$^\star$ \cite{tang_emergent_2023}          &       -&      - &     -  &    -     &   66.0 &   81.1 &   87.2 \\
    ToTF Fuse$^\star$ \cite{zhang_tale_2023}         &   94.3 &   73.2 &   \underline{60.9} &   \underline{79.7}   &   \underline{71.5} &   85.8 &   90.6 \\
    TLfR \cite{zhang_telling_2023}                  &       -&      - &     -  &    -     &   \textbf{74.0} &   \textbf{86.2} &   90.7 \\
    \rowcolor{ourgray}
    DINOv2                                          &   83.9 &   75.1 &   55.1 &    71.4  &   59.4 &   76.8 &   82.4 \\
    \rowcolor{ourgray}
    Ours                                            &   \textbf{98.2} &   \textbf{89.8} &   \textbf{88.8} &   \textbf{92.3}   &   70.6 &   \textbf{86.2} &   \underline{91.1} \\
    \bottomrule
\end{tabular}

%% file: tables_spair_results.tex
\begin{tabular}{l|ccccccccccccccccccc}

    \toprule
    Method                                          & Aero   & Bike   & Bird   & Boat   & Bottle & Bus    & Car    & Cat    & Chair  & Cow    & Dog    & Horse  & Motor  & Person & Plant  & Sheep  & Train  & TV     & All    \\
    \midrule
    \multicolumn{20}{c}{\textsc{per-image pck}~$\alpha_\text{bbox} = 0.1$}\\
    \midrule
    PwarpC \cite{truong_probabilistic_2022}         & -      & -      & -      & -      & -      & -      & -      & -      & -      & -      & -      & -      & -      & -      & -      & -      & -      & -      &   35.3 \\
    CATs \cite{cho_cats_2021}                       &   52.0 &   34.7 &   72.2 &   34.3 &   49.9 &   57.5 &   43.6 &   66.5 &   24.4 &   63.2 &   56.5 &   52.0 &   42.6 &   41.7 &   43.0 &   33.6 &   72.6 &   58.0 &   49.9 \\
    CATs++ \cite{cho_cats_2023}                     &   60.6 &   46.9 &   82.5 &   41.6 &   56.8 &   64.9 &   50.4 &   72.8 &   29.2 &   75.8 &   65.4 &   62.5 &   50.9 &   56.1 &   54.8 &   48.2 &   80.9 &   74.9 &   59.9 \\
    \midrule
    SCOT \cite{liu_semantic_2020}                   &   34.9 &   20.7 &   63.8 &   21.1 &   43.5 &   27.3 &   21.3 &   63.1 &   20.0 &   42.9 &   42.5 &   31.1 &   29.8 &   35.0 &   27.7 &   24.4 &   48.4 &   40.8 &   35.6 \\

    \midrule
    \multicolumn{20}{c}{\textsc{per-keypoint pck}~$\alpha_\text{bbox} = 0.1$}\\
    \midrule
    ASIC \cite{gupta_asic_2023}                     &   57.9 &   25.2 &   68.1 &   24.7 &   35.4 &   28.4 &   30.9 &   54.8 &   21.6 &   45.0 &   47.2 &   39.9 &   26.2 &   48.8 &   14.5 &   24.5 &   49.0 &   24.6 &   36.9 \\
    Dis.Dift$^\dagger$ \cite{fundel_distillation_2025}     & -      & -      & -      & -      & -      & -      & -      & -      & -      & -      & -      & -      & -      & -      & -      & -      & -      & -      &   70.6 \\
    S.Maps \cite{mariotti_improving_2024}          &   74.8 &   64.5 &   87.1 &   45.6 &   52.7 &   77.8 &   71.4 &   82.4 &   47.7 &   82.0 &   67.3 &   73.9 &   67.6 &   60.0 &   49.9 &   69.8 &   78.5 &   59.1 &   67.3 \\
    ToTF \cite{zhang_tale_2023}                    &   81.2 &   66.9 &   91.6 &   61.4 &   57.4 &   85.3 &   83.1 &   90.8 &   54.5 &   88.5 &   75.1 &   80.2 &   71.9 &   77.9 &   60.7 &   68.9 &   92.4 &   65.8 &   74.6 \\
    \midrule
    
    DIFT \cite{tang_emergent_2023}                  &   63.5 &   54.5 &   80.8 &   34.5 &   46.2 &   52.7 &   48.3 &   77.7 &   39.0 &   76.0 &   54.9 &   61.3 &   53.3 &   46.0 &   57.8 &   57.1 &   71.1 &   \textbf{63.4}  &   59.5 \\
    DINOv2 \cite{zhang_tale_2023}                   &   72.7 &   62.0 &   85.2 &   41.3 &   40.4 &   52.3 &   51.5 &   71.1 &   36.2 &   67.1 &   64.6 &   67.6 &   61.0 &   68.2 &   30.7 &   62.0 &   54.3 &   24.2 &   55.6 \\
    SD \cite{zhang_tale_2023}                        &   63.1 &   55.6 &   80.2 &   33.8 &   44.9 &   49.3 &   47.8 &   74.4 &   38.4 &   70.8 &   53.7 &   61.1 &   54.4 &   55.0 &   54.8 &   53.5 &   65.0 &   53.3 &   57.2 \\
    ToTF \cite{zhang_tale_2023}                      &   73.0 &   64.1 &   86.4 &   40.7 &   \underline{52.9} &   55.0 &   53.8 &   78.6 &   45.5 &   77.3 &   64.7 &   69.7 &   \underline{63.3} &   69.2 &   \underline{58.4} &   \underline{67.6} &   66.2 &   53.5 &   64.0 \\
    TLfR$^\dagger$ \cite{zhang_telling_2023}         &   \textbf{78.0} &   \textbf{66.4} &   \textbf{90.2} &            44.5 &   \textbf{60.1} &   \textbf{66.6} &   \textbf{60.8} &   \textbf{82.7} &   \textbf{53.2} &   \textbf{82.3} &   \underline{69.5} &   \textbf{75.1} &   \textbf{66.1} &   \textbf{71.7} &   \textbf{58.9} &   \textbf{71.6} &   \textbf{83.8} &   \underline{55.5} &   \textbf{69.6} \\
    \rowcolor{ourgray}
    DINOv2                                          &   72.5 &   63.3 &   85.7 &   39.9 &   41.4 &   51.6 &   51.7 &   71.0 &   35.8 &   67.9 &   65.5 &   67.9 &   59.2 &   67.2 &   30.1 &   60.8 &   54.7 &   25.0 &   55.7 \\
    \rowcolor{ourgray}
    Ours                                            &  73.6 &   65.1 &   87.6 &   \underline{45.5} &   50.5 &   55.1 &   56.5 &   79.3 &   45.0 &   76.1 &   67.7 &   70.1 &   61.6 &   69.6 &   41.8 &   64.0 &   66.8 &   41.3 &   62.3 \\
    \rowcolor{ourgray}
    Ours$^\dagger$                                  &  \underline{77.9} &   \underline{66.2} &   \underline{88.1} &   \textbf{45.8} &   50.6 &   \underline{63.1} &   \underline{59.6} &   \underline{81.4} &   \underline{48.1} &   \underline{79.1} &   \textbf{69.8} &   \underline{71.8} &   62.2 &   \underline{70.2} &   42.5 &   65.2 &   \underline{79.3} &   41.8 &   \underline{65.3}\\
    \bottomrule
\end{tabular}

%% file: tables_comp_results.tex
\begin{tabular}{lcccccc}
    \toprule
    Method                          & SD            & DINO          & Params    & Feat. Ext. (ms) & Match (ms) & Total (ms) \\
    \midrule
    DIFT\cite{tang_emergent_2023}   & \checkmark    &               & 1.2B      & 2310 & 1      & 2311\\
    TOTF\cite{zhang_tale_2023}      & \checkmark    & \checkmark    & 1.3B      & 1880 & 17     & 1917 \\
    TlFR\cite{zhang_telling_2023}   & \checkmark    & \checkmark    & 1.3B      & 1880 & 4      & 1904 \\
    DINOv2                          &               & \checkmark    & 87M       & 97   & 2      & 99\\
    Ours (GW)                       &               & \checkmark    & 87M       & 97   & 120    & 217\\
    \bottomrule
\end{tabular}

%% file: tables_ablation.tex
\begin{tabular}{ccccccc}
    \toprule
    OT          & GW            & UB            & Sym           &Pose-Align      & PCK($\alpha=0.1$) &\\
    \midrule
                &               &               &               &                & 55.7             &\\
    \checkmark  &               &               &               &                & 55.9             & \textcolor{green}{+0.3} \\
    \checkmark  & \checkmark    & \checkmark    &               &                & 59.2             & \textcolor{green}{+3.3} \\
    \checkmark  & \checkmark    & \checkmark    & \checkmark    &                & 62.3             & \textcolor{green}{+3.1} \\
    \checkmark  & \checkmark    & \checkmark    & \checkmark    &\checkmark      & 65.3             & \textcolor{green}{+2.0} \\
    \bottomrule
\end{tabular}

%% file: tex_7_conclusion.tex
\section{Conclusion}

In this work we present a novel optimal transport based matching algorithm for semantic correspondences. Unlike prior works that rely on Stable Diffusion features to encourage spatial smoothness, our method directly integrates this property into the matching process achieving competitive performance while requiring significantly less compute and memory. The proposed algorithm attains state-of-the-art results on the TSS dataset and competitive results on the PASCAL and SPair datasets. Promising directions for future works include exploring adaptive filter sizes for the Gromov–Wasserstein formulation, extending pairwise penalties to higher-order interactions, and applying our framework to video label propagation tasks.
\bigskip

\noindent\textbf{Acknowledgments.} This work was supported by an Australian Research Council (ARC) Linkage grant (LP21020093).

%% file: tex_8_supplementary.tex
\section{Implementation Detail}

We use the ViT-B/14 DINOv2 model for all experiments with image resolution 840x840, for a total of 60x60 tokens of feature dimension 768 per image. As the TSS and PF-PASCAL datasets benefit more from a strong spatial prior we use a stronger Gromov Wasserstein weight for these datasets compared to SPair-71k. We only apply the symmetry aware loss for the SPair-71k dataset using the symmetric keypoints defined by Zhang~\etal~\cite{zhang_telling_2023}. All hyperparameters are in \cref{tab:exp_hyper_params}.

\begin{table}[h!]
    \centering
    \begin{tabular}{l|cccccc}
        \toprule
        Dataset         & $\lambda$ & $\lambda_\text{GW}$ & $\lambda_\text{sym}$ & $\lambda_\text{ub}$ & $\delta_\text{min}$ & $\delta_\text{max}$ \\
        \midrule
        TSS             & 0.2       & 0.2                 & -                    & 0.05                & 3                   & 5 \\
        PF-PASCAL       & 0.2       & 0.2                 & -                    & 0.05                & 3                   & 5 \\
        SPair-71k       & 0.6       & 0.1                 & 0.1                  & 0.01                & 3                   & 5 \\
        \bottomrule
    \end{tabular}
    \caption{Hyperparameters for our matching algorithm used in \cref{sec:experiments} of the main paper. }
    \label{tab:exp_hyper_params}
\end{table}

\section{Per Category Quantitative Results}
\begin{figure}
    \centering
    \includegraphics[width=0.8\textwidth]{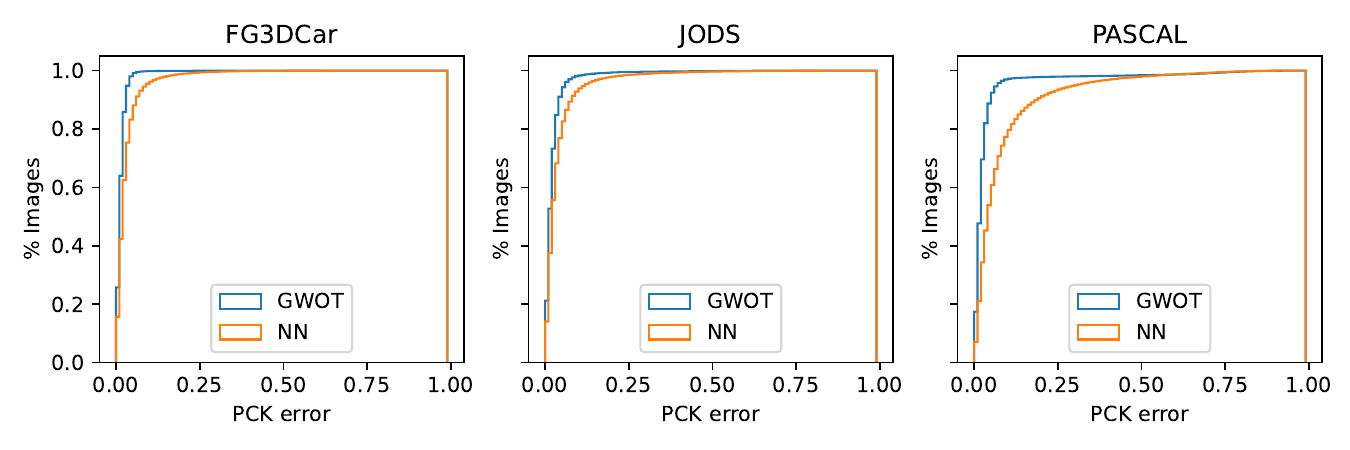}
    \caption{PCK error curves for TSS dataset per category. Our method significantly improves performance compared to the baseline.}
    \label{fig:tss_pck_curves}
\end{figure}

\begin{figure}
    \centering
    \includegraphics[width=\textwidth]{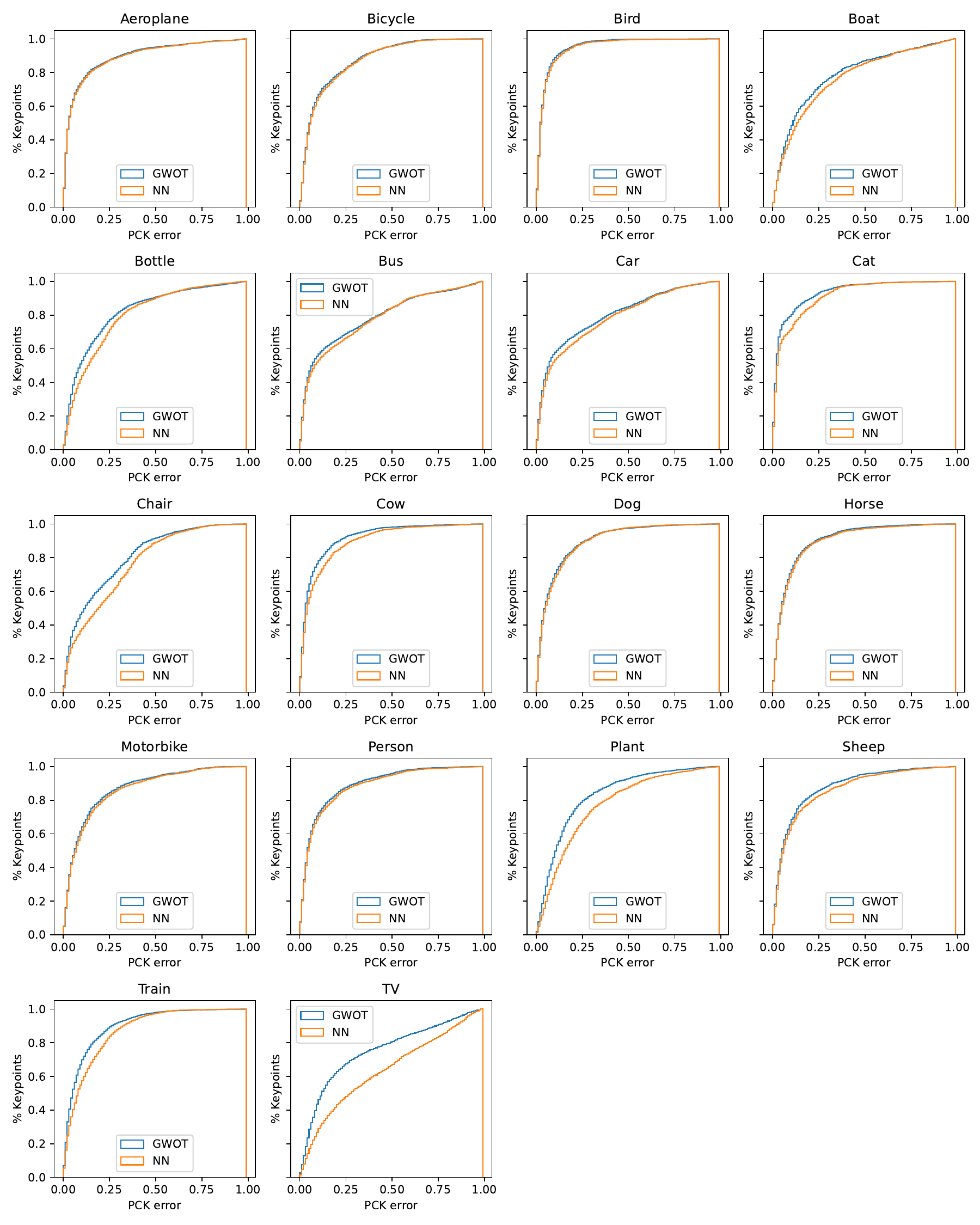}
    \caption{PCK error curves for Spair dataset per category. Our method consistently improves compared to the baseline, in some cases substantially.}
    \label{fig:spair_pck_curves}
\end{figure}

In the main paper we evaluate datasets at set thresholds following standard evaluation protocols. We provide the full curves per category for the Spair-71k and TSS datasets in \cref{fig:spair_pck_curves,fig:tss_pck_curves}. Our method significantly improves on the baseline for all TSS categories. The improvement is more moderate on most Spair-71k categories, with significant improvement on the \texttt{Chair}, \texttt{Plant}, and \texttt{TV} categories.

\section{Qualitative Examples}
We provide qualitative examples of correspondences using our method on the Spair-71k dataset. The background is masked out using ground truth segmentation maps for visualisation. The correspondences are spatially smooth and plausible for a range of object categories including non-rigid categories such as \texttt{Cow} (rows 1 and 2), \texttt{Sheep} (rows 3 and 4) and \texttt{Dog} (rows 11 and 12).
\begin{figure}
    \centering
    \includegraphics[width=0.95\textwidth]{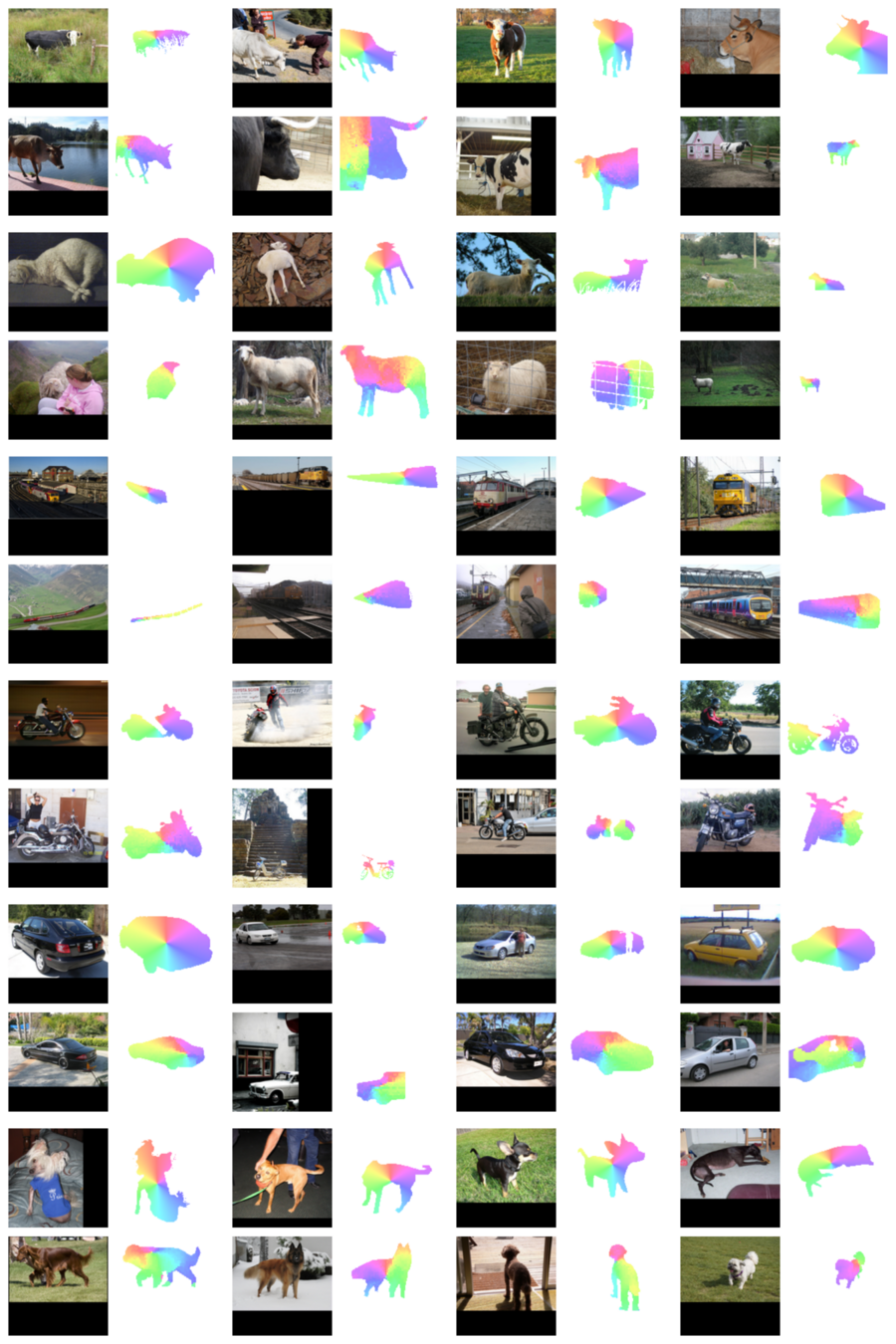}
    \caption{Qualitative examples for different categories in the SPair-71k dataset.}
\end{figure}